\pdfoutput=1
\documentclass[11pt]{article}
\usepackage[final]{acl}
\usepackage{times}
\usepackage{latexsym}
\usepackage{amsmath}
\usepackage[T1]{fontenc}
\usepackage[utf8]{inputenc}
\usepackage{microtype}
\usepackage{inconsolata}
\usepackage{graphicx}
\usepackage{longtable}
\usepackage{multirow}
\usepackage{tabularx, colortbl}
\usepackage{url}
\usepackage{paralist}
\usepackage{makecell}
\usepackage{multirow}
\usepackage{booktabs}
\usepackage{xspace}

\usepackage{annotates}
\newcommand{\F}{F$_1$\xspace}
\newcommand{\emodefabel}{\texttt{Emo-DeFaBel}\xspace}
\newcommand{\defabel}{\texttt{DeFaBel}\xspace}
\newcommand{\gpt}{\texttt{GPT}\xspace}
\newcommand{\falcon}{\texttt{Falcon}\xspace}
\newcommand{\llama}{\texttt{Llama}\xspace}

\newcommand{\emohi}[1]{\textsc{#1}}
\newcommand{\anger}{\emohi{anger}\xspace}
\newcommand{\joy}{\emohi{joy}\xspace}
\newcommand{\fear}{\emohi{fear}\xspace}
\newcommand{\sadness}{\emohi{sadness}\xspace}
\newcommand{\disgust}{\emohi{disgust}\xspace}
\newcommand{\surprise}{\emohi{surprise}\xspace}
\newcommand{\pride}{\emohi{pride}\xspace}
\newcommand{\shame}{\emohi{shame}\xspace}
\newcommand{\guilt}{\emohi{guilt}\xspace}
\newcommand{\interest}{\emohi{interest}\xspace}
\newcommand{\noemotion}{\emohi{no emotion}\xspace}

\makeatletter
\renewcommand{\paragraph}{%
  \@startsection{paragraph}{4}%
  {\z@}{0.75ex \@plus 1ex \@minus .2ex}{-1em}%
  {\normalfont\normalsize\bfseries}%
}
\makeatother

\title{Fearful Falcons and Angry Llamas:\\Emotion Category Annotations of Arguments by Humans and LLMs}

\author{
  Lynn Greschner \and Roman Klinger\\
  Fundamentals of Natural Language Processing, University of Bamberg, Germany\\
  \texttt{\{lynn.greschner,roman.klinger\}@uni-bamberg.de}\\
}

\begin{document}
\maketitle
\begin{abstract}
  Arguments evoke emotions, influencing the effect of the argument
  itself. Not only the emotional intensity but also the category
  influence the argument's effects, for instance, the willingness to
  adapt stances.  While binary emotionality has been studied in
  arguments, there is no work on discrete emotion categories (e.g., `Anger')
  in such data. To fill this gap, we crowdsource subjective
  annotations of emotion categories in a German argument corpus
  and evaluate automatic LLM-based labeling methods.  Specifically, we
  compare three prompting strategies (zero-shot, one-shot,
  chain-of-thought) on three large instruction-tuned language models
  ({Falcon-7b-instruct}, {Llama-3.1-8B-instruct}, {GPT-4o-mini}). We
  further vary the definition of the output space to be binary (is
  there emotionality in the argument?), closed-domain (which emotion
  from a given label set is in the argument?), or open-domain (which
  emotion is in the argument?).
  We find that emotion categories enhance the prediction of emotionality in arguments, emphasizing the need for discrete emotion annotations in arguments. Across all prompt
  settings and models, automatic predictions show a high recall
  but low precision for predicting anger and fear,
  indicating a strong bias toward negative emotions.
\end{abstract}

\section{Introduction}
The role of emotionality received considerable attention in recent
research, spanning from a focus on pathos
\citep{evgrafova-etal-2024-analysing} to the study of emotion
intensity and its role on argument persuasiveness
\citep{benlamine-2017-persuasive-argumentation,griskevicius_influence_2010}.
In natural language processing, most research focused on binary or
continuous emotion concepts \citep[i.a.]{el-baff-etal-2020-analyzing},
and the role of such variables on argument effectiveness has been
confirmed in empirical studies. In argumentation theory and
psychology, however, it has also been shown that not only binary
emotionality or intensity are important, but also the concrete emotion
category, or groups of emotions, matter. Positive emotions, for instance
have a positive effect on cognitive abilities and therefore the
willingness to adapt an own stance
\citep{griskevicius_influence_2010}. Negative emotions are often part
of persuasion strategies \citep{boster_meta-analytic_2016} while they
can also lead to defense behaviour \citep{leventhal_negative_1968}.
We therefore argue that there is a mismatch between research in
natural language processing for argument mining and psychological and
theoretical work on argument analysis.

\begin{table}[t]
  \centering\small
  \begin{tabularx}{\linewidth}{Xp{8mm}}
    \toprule
    Argument & Label \\
    \cmidrule(r){1-1}\cmidrule(lr){2-2}
    Es gibt Impfstoffe, welche unsere DNA dauerhaft verändern können. Diese Impfstoffe nennen sich mRNA-Impfstoffe. Das mRNA gelangt in unsere DNA und gliedert sich dort mit ein. Dadurch wird unsere DNA leicht verändert.\par \textit{There are vaccines that can permanently alter our DNA. These vaccines are called mRNA vaccines. The mRNA gets into our DNA and integrates into it. This slightly alters our DNA.}  & Interest,\par Disgust,\par Surprise\\
    \bottomrule
  \end{tabularx}
  \caption{Example argument with discrete emotion labels from human annotators in \emodefabel (English translation in italics).}
  \label{tab:intro_example}
\end{table}

To fill this gap, we approach argumentative texts as a domain for
emotion analysis, more concretely the most prominent subtask of
emotion categorization. In this research direction, the goal is to
assign emotion concepts to predefined textual units, for instance
literary texts \citep{mohammad-2011-upon}, political debates \citep{tarkka-etal-2024-automated}, or social media texts \citep{mohammad-etal-2014-semantic}. Emotions
are often expressed implicitly, without explicitly mentioning emotion
concepts
\citep{casel-etal-2021-emotion,klinger-etal-2018-iest,koga-etal-2024-forecasting-implicit,lee-lau-2020-event}. That
renders emotion classification a challenging task, even for humans,
who tend to agree more with other readers of an emotional text than
with the original experiencer of the emotion
\citep{troiano-etal-2023-dimensional}.

\begin{table*}
\centering\small
    \centering
    \begin{tabularx}{\linewidth}{p{3cm}X}
      \toprule
      Statement & Argument  \\
      \cmidrule(r){1-1}\cmidrule(l){2-2}           
      Kamele speichern
      Fett in ihren Höckern. &
                               Kamele sind sehr große Tiere und benötigen sehr viel Energie. Um
                               diese Energie aus den Fettreserven zu erhalten, wird das Fett in den
                               Höckern gespeichert. Da Kamele sich meistens in Gegenden befinden,
                               in denen sie wenig Nahrung finde und dort als Lastentiere eingesetzt
                               und lange Wege zurücklegen, werden, ist es überaus wichtig, sich
                               vorher einen Fettspeicher anzulegen. Außerdem schützen die mit Fett
                               gefüllten Höcker die Kamele vor der Hitze und der Sonne, sie wirken
                               wie eine Art Polster, dass die übrigen Organe vor Überhitzung schützt.\\
      \textit{Camels store fat in their hump.} &
                                                 \textit{Camels are very large animals and need a lot of energy. To
                                                 energy from the fat reserves, the fat is stored in the humps.
                                                 humps. Since camels are usually found in areas
                                                 where they find little food and are used there as beasts of burden
                                                 and travel long distances, it is extremely important to build up a fat
                                                 to build up a fat store beforehand. In addition, the fat-filled humps
                                                 humps filled with fat protect the camels from the heat and the sun, they act
                                                 They act like a kind of cushion that protects the other organs from overheating.} \\
      \bottomrule
    \end{tabularx}
    \caption{Example statement and participant-generated argument from the \defabel corpus \citep{velutharambath-etal-2024-factual}. English translation in italics.}
    \label{tab:defabel_data_example}
\end{table*}

Hence, we crowdsource human emotion and convincingness annotations for
the publicly available German DeFaBel corpus
\citep{velutharambath-etal-2024-factual} and compare them to
automatically assigned labels by three large language models in three
prompting approaches. Our main contribution is therefore the corpus
\emodefabel, (1) the first argumentative corpus human-labeled with
emotion categories and (2), an analysis of the performance of the
language models for emotion analysis. Analyzing \emodefabel reveals that joy and pride in arguments are correlated with
higher convincingness, while anger is negatively correlated. Our experiments demonstrate that the prompt-based
model categorizations are heavily biased toward negative emotions. The biases on concrete emotions differ between the models.

\section{Related Work}
\subsection{Language Models for Emotion Analysis}
With the rise of LLMs, utilizing such models for emotion analysis has
received some attention. \citet{churina-etal-2024-wassa} explored the
capabilities of LLMs for empathy and emotion prediction in
dialogues. \citet{cheng-etal-2024-teii, nedilko-2023-generative}
focused on multilingual analyses. \citet{bagdon-etal-2024-expert}
studied best--worst scaling as an approach for emotion
intensity annotations with language models.

 Generally, LLMs may replicate human annotations well, for instance in
 Finnish parliamentary debates \citep[using
 GPT4,][]{tarkka-etal-2024-automated}.
 \citet{malik_multilabel_emotion_annotation} report a similar success
 for French Tweets.  \citet{Gilardi2023ChatGPT_emotion_annotation}
 highlight performance and cost advantages of such approach over
 manual annotations. We follow this prior work and transfer it to
 emotion analysis for argument data.

 \subsection{Prompting Approaches}
 One important challenge when prompting language models for language
 understanding tasks is to find well-performing instructions
 \citep{ye-etal-2024-prompt}.  Despite efforts to automatically create
 appropriate prompts \citep{li-etal-2023-robust,
   chen-etal-2024-prompt}, prompts
 commonly need to be adapted to a domain at hand, and are not
 sufficiently robust across use-cases.

\citet{Reynolds2021PromptPF} demonstrate that zero-shot prompts can
outperform one-shot prompts, arguing that providing examples does not
necessarily improve performance. \citet{fonseca-cohen-2024-large}
explore LLMs learning capabilities for new facts or concept
definitions through prompts. Their results find that zero-shot
prompting improves sentence labeling performance, but larger models
(70B+ parameters) struggle with counterfactual scenarios. GPT-3.5 was
the only model to detect nonsensical guidelines, while
Llama-2-70B-chat often outperformed Falcon-180B-chat, suggesting that
increasing model size alone does not guarantee better adherence to
guidelines.

Numerous experimental results suggest that chain-of-thought prompting
leads to performance improvements
\citep[i.a.]{kojima_2022_llms_zero-shot, du-etal-2023-task}. In
contrast, \citet{le-scao-rush-2021-many} find that the prompt choice
is not the most dominant parameter when optimizing model performance
in low-data regimes. We, therefore, consider three commonly used
prompting approaches (zero-shot, one-shot, and chain-of-thought
approaches) for emotion analysis in arguments.

\begin{table*}[t]
  \centering\sffamily\small\scalefont{0.85}
  \renewcommand{\arraystretch}{0.7}
  
  \begin{tabularx}{\linewidth}{p{2mm}p{7mm}XXX}
 
    \toprule
    && Binary & Closed-domain & Open-domain \\
    \cmidrule(lr){3-3}\cmidrule(lr){4-4}\cmidrule(l){5-5}

    \multirow{4}{*}{\rotatebox[origin=c]{90}{Zero-shot}} 
    & Role 
    & \multicolumn{3}{p{13.84cm}}{\centering\cellcolor{orange!30}You are an expert on emotions in arguments.} \\

    \cmidrule(lr){2-2}\cmidrule(lr){3-3}\cmidrule(lr){4-4}\cmidrule(l){5-5}
    & Task Desc.
    & \cellcolor{yellow!10} Label the following argumentative text about a statement into containing emotion(s) (emotion:1) or not containing emotions (emotion:0). 
    & \cellcolor{red!10} Your task is to label an argumentative text about a statement with the most present emotion a reader would feel. The options for labels are:  [Emo].  
    & \cellcolor{blue!10} Your task is to label an argumentative text about a statement with the most present emotion a reader would feel. \\ 
    \cmidrule(lr){2-2}\cmidrule(lr){3-3}\cmidrule(lr){4-4}\cmidrule(l){5-5}
    & Format
    & \multicolumn{3}{p{\dimexpr\linewidth-21.6mm}}{\cellcolor{orange!30}Provide the output in a json format with the key being 'emotion' and the value being the emotion label as a string. For example, if you believe the argument contains sadness, your json output should be:} \\
    \cmidrule(lr){3-3}\cmidrule(lr){4-4}\cmidrule(l){5-5}
    & & \cellcolor{yellow!10}`1'. & \multicolumn{2}{c}{\cellcolor{red!50!blue!20}`sadness'.} \\
    \cmidrule(lr){2-2}\cmidrule(lr){3-3}\cmidrule(lr){4-4}\cmidrule(l){5-5}
    &Texts& \multicolumn{3}{p{\dimexpr\linewidth-21.6mm}}{\cellcolor{orange!30}Now label the following argumentative text with the emotion label. Statement: \{statement\}. Text: \{text\}. What is the emotion label for this argument? Only output the json format.} \\
    \cmidrule(r){1-2}\cmidrule(lr){3-3}\cmidrule(lr){4-4}\cmidrule(l){5-5}

    \multirow{5}{*}{\rotatebox{90}{One-shot}}&Role & \multicolumn{3}{c}{\cellcolor{orange!30}You are an expert on emotions in arguments.} \\
    \cmidrule(lr){2-2}\cmidrule(lr){3-3}\cmidrule(lr){4-4}\cmidrule(l){5-5}
    & Task Desc.&  \cellcolor{yellow!10}Label the following argumentative text about a statement into containing emotion(s) (emotion:1) or not containing emotions (emotion:0). 
    & \cellcolor{red!10} Your task is to label an argumentative text about a statement with the most present emotion a reader would feel. The options for labels are:  [Emo].  
    & \cellcolor{blue!10} Your task is to label an argumentative text about a statement with the most present emotion a reader would feel. \\ 
    \cmidrule(lr){2-2}\cmidrule(lr){3-3}\cmidrule(lr){4-4}\cmidrule(l){5-5}
    & Ex. & \multicolumn{3}{c}{\cellcolor{orange!30}[Example with correct output.]}\\
    \cmidrule(lr){2-2}\cmidrule(lr){3-3}\cmidrule(lr){4-4}\cmidrule(l){5-5}
    & Format & \multicolumn{3}{p{\dimexpr\linewidth-21.6mm}}{\cellcolor{orange!30}Provide the output in a json format with the key being 'emotion' and the value being the emotion label as a string. For example, if you believe the argument contains `sadness', your json output should be:} \\
    \cmidrule(lr){3-3}\cmidrule(l){4-4}\cmidrule(l){5-5}
    & & \cellcolor{yellow!10}`1'. & \multicolumn{2}{c}{\cellcolor{red!50!blue!20}`sadness'.} \\
    \cmidrule(lr){2-2}\cmidrule(lr){3-3}\cmidrule(lr){4-4}\cmidrule(l){5-5}
    &Texts& \multicolumn{3}{p{\dimexpr\linewidth-21.6mm}}{\cellcolor{orange!30} Now label the following argumentative text with the emotion label. Statement: \{statement\}. Text: \{text\}. What is the emotion label for this argument? Only output the json format.} \\
    \cmidrule(r){1-2}\cmidrule(lr){3-3}\cmidrule(l){4-4}\cmidrule(l){5-5}

    \multirow{5}{*}{\rotatebox{90}{Chain-of-Thought}}&Role &  & \multicolumn{2}{c}{\cellcolor{red!50!blue!20}You are an expert on emotions in arguments.} \\
    \cmidrule(r){2-2}\cmidrule(l){4-4}\cmidrule(l){5-5} 
    & Task Desc.&   & \cellcolor{red!10} Your task is to first classify an argumentative text into either containing an 
    emotion or not containing an emotion. If the text contains an emotion, continue with the following task: Your task is to label the text with one emotion a reader would feel the strongest when reading the argument. 
    The option for labels are: [Emo]
    In your answer, only provide the emotion label you choose as the output. & \cellcolor{blue!10} Your task is to first classify an argumentative text into either containing an 
    emotion or not containing an emotion. If the text contains an emotion, continue with the following task: Your task is to label the text with one emotion a reader would feel the strongest when reading the argument. 
    In your answer, only provide one emotion label you choose as the output. \\

    \cmidrule(lr){2-2}\cmidrule(l){4-4}\cmidrule(l){5-5}
    & Ex.&  &  \multicolumn{2}{c}{\cellcolor{red!50!blue!20}[Example with correct output.]} \\
    \cmidrule(lr){2-2}\cmidrule(l){4-4}\cmidrule(l){5-5}

    &Format&  & \multicolumn{2}{p{9.1cm}}{\cellcolor{red!50!blue!20}Provide the output in a json format with the key being 
    'emotion' and the value being the emotion label as a string. For example, if you believe the argument
     contains fear, your json output should be: \{'emotion': Fear\}". } \\
    \cmidrule(lr){2-2}\cmidrule(l){4-4}\cmidrule(l){5-5}
    &Texts&  & \multicolumn{2}{p{9.1cm}}{\cellcolor{red!50!blue!20} Now label the following argumentative text with the emotion label. Statement: \{statement\}. Text: \{text\}. What is the emotion label for this argument? Only output the json format.} \\

    \bottomrule
  \end{tabularx}
  \caption{Prompt templates for emotion domain (binary, closed-domain,
    open-domain) and prompting settings (zero-shot, one-shot,
    chain-of-thought). [Emo] refers to the set of \joy, \anger, \fear,
    \sadness, \disgust, \surprise, \pride, \interest, \shame, \guilt,
    \noemotion. [Example with correct output] consists of a
    human-annotated argument with an emotion label and is the
    consistent for all prompts. Color highlights shared elements.}
  \label{tab:prompt_templates}
\end{table*}

\subsection{Emotions in Arguments}
\citet{habernal-gurevych-2016-argument,
  habernal-gurevych-2017-argumentation} constructed and analyzed a
corpus for convincingness strategies in online argumentative text,
including emotionality.  \citet{lukin-etal-2017-argument} highlight the role
of emotions in interaction with differing personalities on the
perceived convincingness of arguments.  There is further a substantial
body of psychological studies which point to the role of cognitive
argument evaluations for convincingness
\citep{bohner_affect-persuasion-1992, petty_mood_persuasion_1993,
  pfau_affect_resistance_2006, Worth1987CognitiveMO,
  Benlamine2015EmotionsArgumentesEmpirical}. \citet{benlamine_2017-persuation-emotions}
demonstrate that the argumentation strategy \textit{Pathos} (i.e.,
using emotions) is most efficient for changing a persons opinion.

Research in NLP focuses, so far, on binary emotionality and emotion
intensity in arguments, as one factor of convincingness of many
\citep{habernal-gurevych-2017-argumentation} or rate the emotional
appeal \citep{wachsmuth-etal-2017-computational,
  lukin-etal-2017-argument}. We are not aware of work that focuses on
discrete emotion categories. Our study closes this gap.

\section{Annotation}
\label{sec:annotation}
We need emotion labels of argumentative texts in our study and
therefore enhance an existing argument dataset with emotion labels. In
total, we request three individual annotations for 300 arguments.

\paragraph{Data Sets.}
The basis for our annotation is the \defabel corpus
\citep{velutharambath-etal-2024-factual}. It contains argumentative
German texts, annotated via crowdsourcing. The participants were asked
to write persuasive arguments supporting a given statement (e.g.,
``Camels store fat in their hump.''), selected from the TruthfulQA
dataset \citep{lin-etal-2022-truthfulqa}. The corpus contains 1031
arguments for 35 statements. We select this resource because it
contains short, isolated arguments. Additionally, this dataset allows
us to effortlessly capture the annotators' stances toward each
statement. We randomly select 300 arguments, evenly distributed across
statements for our annotation
task. Table~\ref{tab:defabel_data_example} shows an example
statement--argument pair.

\paragraph{Annotation Setup.}
\label{sec:annotation_setup}

\begin{table}[t]
\centering\small
    \centering
    \begin{tabularx}{\linewidth}{p{7mm}Xc}
    \toprule
    Var. & Question Text & Label \\
    \cmidrule(r){1-1}\cmidrule(lr){2-2}\cmidrule(rl){3-3}
        Stance & Stimmen Sie der Aussage zu?\par \textit{Do you agree with the statement?} & 1--5\\
        Fam. & Wie gut kennen Sie sich mit dem Thema aus?\par \textit{How familiar are you with the topic?} & 1--5\\
        Conv. & Wie überzeugend ist das Argument für Sie?\par \textit{How convincing is this argument for you?} & 1--5 \\
        Binary & Wird eine Emotion in Ihnen ausgelöst wenn sie das Argument lesen?\par \textit{Is an emotion triggered in you when you read the argument?} & Yes/No \\
        Emo. & Beantworten Sie diese Frage nur wenn Sie die vorangegangene Frage mit ``Ja'' beantwortet haben. Welche der folgenden Emotionen wird am stärksten in Ihnen ausgelöst wenn Sie das Argument lesen?\par \textit{Only answer this question if you have answered the previous question with ``Yes''. Which of the following emotions is triggered most strongly in you when you read the argument?} & [Emo] \\
    \bottomrule
    \end{tabularx}
    \caption{Wording and response options for the human annotation study. [Emo] refers to Freude, Wut, Angst, Traurigkeit, Ekel, Überraschung, Stolz, Interesse, Scham, Schuld (\textit{Joy , Anger, Fear, Sadness, Disgust, Surprise, Pride, Interest, Shame, Guilt}). We conducted the study in German, see English translations in italics.}
    \label{tab:annotation_study_design}
\end{table}

We show one statement--argument pair per page. The participants are
instructed to read those texts and provide their stance toward the statement
on a 5-point scale (strongly agree, \ldots, strongly disagree). In
addition, we ask how convincing they perceive the argument on a
5-point scale (not convincing at all, \ldots, very convincing). For
the emotion label, we first ask if the argument evokes an emotion in
the participants (yes/no). If they answer yes, they are asked to
provide the concrete emotion label from our emotion label set (\joy ,
\anger, \fear, \sadness, \disgust, \surprise, \pride, \interest,
\shame, \guilt). Further, we introduce a free text field to input an
emotion that is not covered by the label set, which allows us a more
fine-grained analysis of the subjective discrete emotion labels in
arguments. Table \ref{tab:annotation_study_design} displays an
overview of the collected labels, question phrasings, and possible
answers. We show a screenshot of one example annotation page in the
Appendix in Figure~\ref{fig:screenshot_1}. The free-text field answers are
collapsed to a closed set for further modeling and analysis (see
Appendix~\ref{tab:emotion_mapping_appendix}).

\paragraph{Crowd-sourcing Details.}
We use the platform Prolific\footnote{\url{https://www.prolific.com/}}
with Potato \citep{pei-etal-2022-potato}. Participants are filtered to
be in Germany, have German as their first and native language, be
fluent in German, and have an approval rate of 80--100\%.

Each participant answers the survey for five statement--argument pairs
(and one attention check, see Figure~\ref{fig:attention_check} for an example). We pay each participant 1.20\texteuro{}
for one survey, which on average takes 7 minutes. Participants can
participate up to 12 times, and therefore annotate up to 60
statement--argument pairs. In total, the cost of the study amounts to
316.87\texteuro. The contributing participants of our studies were on
average 35.8 years old (21 minimum, 68 maximum). From this set, 98
identified as male and 96 as female (we did not limit the study to
these two genders).

\section{Models}

To investigate the efficiency of LLM annotation of emotions in
arguments we differentiate between two dimensions, the emotion domain
setting (binary, closed-domain, open-domain) and the technique
(zero-shot, one-shot, and chain-of-thought) of the prompts. Combining
these strategies results in eight prompt formulations (cf.\
Table~\ref{tab:prompt_templates}).\footnote{The supplementary material
  for this paper (code and annotated data) can be found here: \url{https://www.uni-bamberg.de/en/nlproc/resources/emodefabel/}.}

\subsection{Emotion Domain}
We differentiate between prompting for emotionality (binary) and for discrete emotion categories (closed and open-domain) in arguments. 

\paragraph{Binary.}
In the binary prompt setting, we request a label for the argument
indicating if it causes an emotion in the reader or not. We do not
distinguish concrete emotion categories. This setting enables us to
develop an understanding regarding an agreement without focusing on
specific categories.

\paragraph{Closed-domain.}
The binary setting is contrasted with the request for concrete emotion
labels, to enable more detailed analysis of specific categories. We
use the label set of \joy , \anger, \fear, \sadness, \disgust,
\surprise, \pride, \interest, \shame, \guilt, or \noemotion, as a
combination of basic emotions, cognitive evaluations (interest) and
self-directed states (shame, guilt, pride). We consider the latter to
be particularly relevant for emotion analysis in argumentative texts.

\paragraph{Open-domain.}
Hypothetically, a language model may perform well to assign emotion
names different from our set. To evaluate this observation in an
open-domain setting in which we do not predefine the emotion set. The
goal of this approach is to capture a broad range of emotions.

\subsection{Prompting Approach}
To study the impact of the three domain settings mentioned above
across multiple prompting approaches, we design three types of
prompts, shown in Table~\ref{tab:prompt_templates}.

\paragraph{Zero-shot.}
Zero-shot (ZS) prompts only contain instructions to
complete a given task without any examples or further context. ZS
prompts are flexible, comparably straight-forward to design, and no
examples are required. 
 
\paragraph{One-shot.}
One-shot (OS) prompts augment the instruction to the model with one or few
training examples, allowing the model to learn in context
\citep{brown-2020-LMs-few-shot-learners}.  We perform OS
prompting with a manually annotated example from the \defabel corpus
that we use for our experiments which is not part of our test set.

\paragraph{Chain-of-Thought.}
Since the binary emotionality of a given argument is conditional for
the discrete emotion label, we hypothesize that first creating a
rationale before giving the prediction enhances the performance of
LLMs. Chain-of-thought (CoT) prompting is a technique that triggers the
model to generate a series of logical reasoning steps before providing
the final answer \citep{wei_cot-prompting-neurpis}. This technique
assists models to tackle more complex tasks more effectively by
simulating a human-like reasoning process. In our study, we formulate
the prompt to force the model to first decide on the binary
emotionality of a given argument before providing the discrete emotion
label (therefore, no binary CoT prompt). See Table~\ref{tab:prompt_templates} for concrete examples.

\subsection{Evaluation}
In the following, we explain our evaluation procedure, in which the
LLM-based predictions\footnote{LLM output parsing in Appendix~\ref{sec:appendix_label_extraction}.} are compared to human annotations (which will be
discussed in Section~\ref{sec:annotation}).
We use two different strategies for the evaluation, \textit{relaxed}
and \textit{strict}, to account for the subjectivity of the task.
In the \textit{strict} mode, we compare the model's output to the
majority vote from the human annotations. If there is no majority, we
assign \noemotion.
In the \textit{relaxed} mode, we count an output as true positive if
it matches \textit{any of} the labels provided by the human
annotators. The motivation for this approach is to consider everything
to be a correct output that may be relevant for ``somebody'',
acknowledging the subjective nature of the task. When evaluating the
models' performance for individual emotion classes, we
distribute one count of a false negative prediction across the set of
gold labels.

\begin{figure}[t]
    \centering
    \includegraphics[width=\columnwidth]{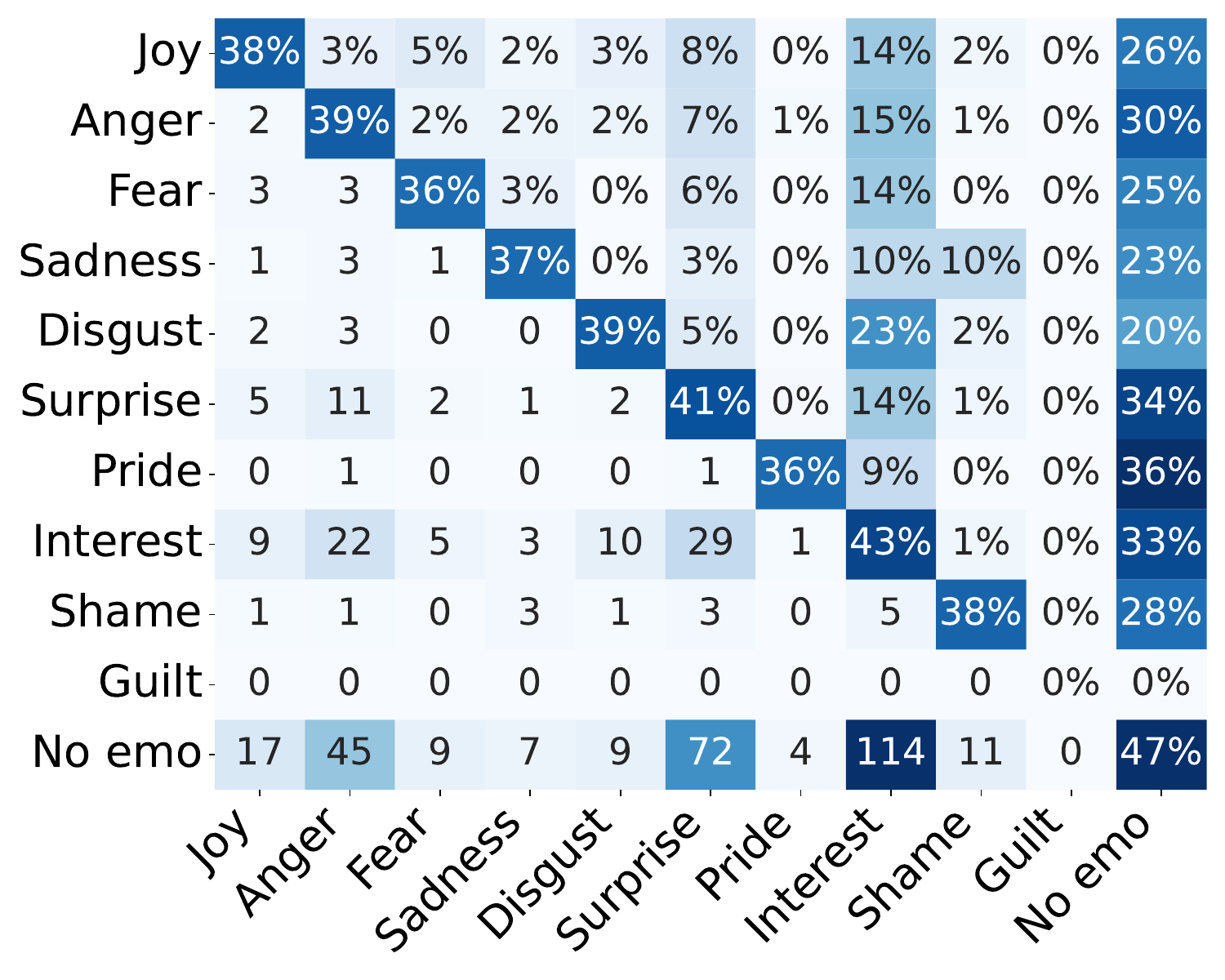}
    \caption{Co-occurences of emotion labels in the human annotation
      study for 300 arguments. The upper part displays percentages, the lower absolute numbers.}
    \label{fig:heatmap_emotions}
\end{figure}

\section{Experiments}
We now explain the experiments, analyze the human annotations and
subsequently answer the research questions stated in the introduction.

\subsection{Experimental Setting}
We use \texttt{
Falcon-7b-instruct} \citep{almazrouei2023falconseriesopenlanguage}, \texttt{
Llama-3.1-8B-Instruct} \citep{dubey2024llama3herdmodels}, and
\texttt{GPT-4o-mini}
\citep{openai2024gpt4technicalreport}\footnote{We refer to
  the models as \falcon, \llama, and \gpt.}. \falcon is an
instruction-tuned generative model with 7 billion parameters. \llama
has 8 billion parameters and is optimized for multilingual dialogue
use cases. We access both models via their respective Huggingface
APIs\footnote{\url{https://huggingface.co/meta-llama/Llama-3.1-8B-Instruct},\url{https://huggingface.co/tiiuae/falcon-7b-instruct}}. With
\texttt{GPT-4o-mini} OpenAI offers a smaller and more cost-efficient
model than GPT-4o that outperforms GPT-3.5-Turbo. We access the model
via the OpenAI API
\footnote{\url{https://openai.com/index/gpt-4o-mini-advancing-cost-efficient-intelligence/}}. We use the default settings for all models. The
cost for \gpt amounts to 0.20\texteuro.

\begin{table}[t]
  \centering
  \small
\begin{tabular}{lrrr}
  \toprule
  & \multicolumn{3}{c}{Num.\ agreem.} \\
  \cmidrule(lr){2-4}
 Emotion & $=$1 & $\leq$2 & $\leq$3\\
\cmidrule(r){1-1}\cmidrule(lr){2-2}\cmidrule(lr){3-3}\cmidrule(lr){4-4}
\joy  & .96 & .04 & .00  \\
\anger & .97 & .03 & .00   \\
\fear& 100  & .00  & .00  \\
\sadness & .91 & .09 & .00  \\
\disgust & .94 & .06 & .00  \\
\surprise & .84 & .15 & .01  \\
\pride & 100  & .00  & .00  \\
\interest & .68 & .28 & .05  \\
\shame & .93 & .07 & .00  \\
\guilt & .00  & .00  & .00  \\
\noemotion & .43 & .43 & .15 \\
\bottomrule
\end{tabular}
\caption{Percentages of emotion label agreements for one, two, and three
  annotators for 300 arguments.}
\label{tab:emotion_agreement_human_study}
\end{table}

\subsection{Results}
\subsubsection{Human Label Analysis}
The human study results in 326 annotations of statement--argument
pairs, from which we keep a random set of 300 instances. Altogether, 16 annotations are rejected due to
failed attention checks.

Out of all arguments, 50\% contain a binary emotion label
(majority-aggregated). The average argument length between emotional
and non-emotional arguments does not differ substantially (78.5
tokens, 4.9 sentences vs.\ 78.4 tokens, 4.7 sentences).  The most
frequently annotated emotion label in the closed-domain annotation is
\interest (207 annotations), followed by \surprise (103). \pride is
the least frequently annotated emotion label (4). Notably, \guilt is
never annotated.

\paragraph{Co-occurences of Emotion Labels.}\label{sec:co-occurences_emo_labels}
The heatmap in Figure~\ref{fig:heatmap_emotions} shows how frequent pair-wise emotion label
co-occurrences are, both in absolute and relative numbers. An
interesting observation is that negative
emotions, such as anger and fear, appear together with the cognitive
emotion interest frequently (15 and 14\%, respectively). In 23\% of
cases, disgust appears together with interest, showcasing the
subjective nature of the discrete emotion labeling task. We speculate
that different emotions might either be evoked from the content of the
argument itself, or be dependent of the annotator's stance toward the
statement, which we discuss further in Section~\ref{sec:qual_analysis}.

\paragraph{Agreement.}
Table~\ref{tab:emotion_agreement_human_study} displays the
inter-annotator agreements, quantified as percentages of agreement of
a given emotion label for one, two, and three annotations of a given
argument. The agreement for an argument containing \fear, \sadness,
and \pride is low; for \surprise and \interest, two annotators agree
in 13\% and 29\% of all arguments. Notably, these labels are also the
most prevalent within the dataset. The highest agreement for all three
annotators agreeing on one label is found for \noemotion. Overall, the
agreement for all emotion labels remains consistently low,
underscoring the inherent subjectivity of emotion annotation
tasks. This variability in agreement is expected given the subtle and
subjective nature of the emotional interpretation of arguments.

\begin{figure}[t]
    \centering
    \includegraphics[width=\columnwidth]{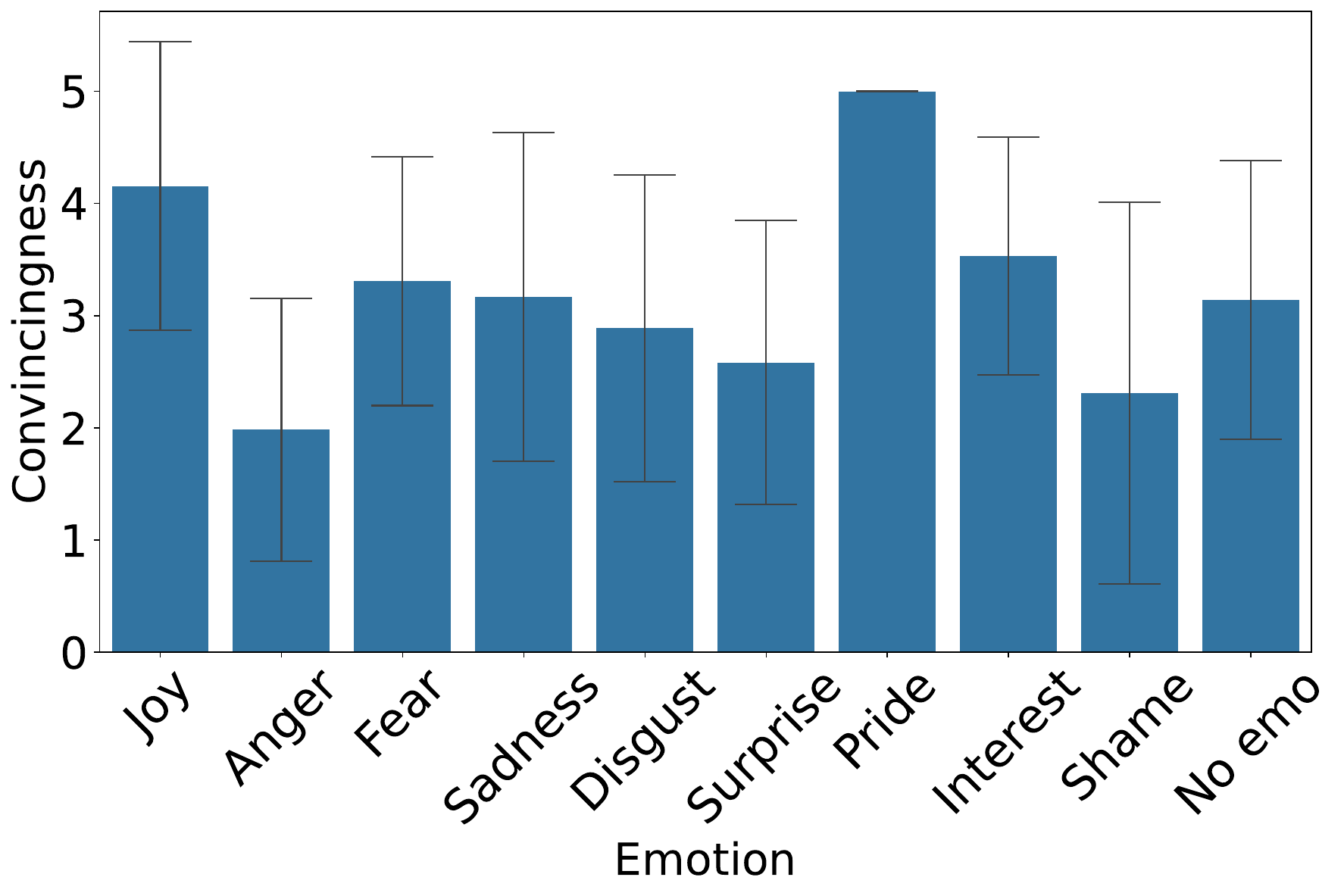}
    \caption{Average convincingness scores (1--5; 5:
      very convincing; 1: not convincing at all) for each
      emotion with standard deviation.}
    \label{fig:barplot_emo_conv}
\end{figure}

\begin{table}[tpb]
  \small
  \centering
  \setlength{\tabcolsep}{3pt} 
\begin{tabular}{ll rrr rrr rrr}
\toprule
& & \multicolumn{3}{c}{\falcon} & \multicolumn{3}{c}{\llama} & \multicolumn{3}{c}{\gpt} \\ 
\cmidrule(r){3-5}\cmidrule(lr){6-8}\cmidrule(l){9-11}
& & \multicolumn{1}{c}{P} & \multicolumn{1}{c}{R} & \multicolumn{1}{c}{\F}& \multicolumn{1}{c}{P} & \multicolumn{1}{c}{R} & \multicolumn{1}{c}{\F}& \multicolumn{1}{c}{P} & \multicolumn{1}{c}{R} & \multicolumn{1}{c}{\F} \\
\cmidrule(r){1-2}\cmidrule(rl){3-3}\cmidrule(lr){4-4}\cmidrule(r){5-5}\cmidrule(l){6-6}\cmidrule(lr){7-7}\cmidrule(r){8-8}\cmidrule(l){9-9}\cmidrule(l){10-10}\cmidrule(l){11-11}
\multirow{2}{*}{\rotatebox{90}{Bin.}}
& ZS & .51 & 1.00 & .67 & .55 & .67 & .61 & .59 & .21 & .31 \\
& OS & .51 & .95 & .66 & .62 & .03 & .06 & .64 & .17 & .26 \\
\cmidrule(r){1-2}\cmidrule(rl){3-3}\cmidrule(lr){4-4}\cmidrule(r){5-5}\cmidrule(l){6-6}\cmidrule(lr){7-7}\cmidrule(r){8-8}\cmidrule(l){9-9}\cmidrule(l){10-10}\cmidrule(l){11-11}
\multirow{3}{*}{\rotatebox{90}{Closed}}
& ZS & .50 & 1.00 & .67 & .51 & .97 & .67 & .51 & .92 & .65 \\
& OS & .51 & .98 & .67 & .50 & 1.00 & .67 & .50 & .94 & .65 \\
& CoT & .50 & .97 & .66 & .50 & 1.00 & .67 & .54 & .71 & .61 \\
\cmidrule(r){1-2}\cmidrule(rl){3-3}\cmidrule(lr){4-4}\cmidrule(r){5-5}\cmidrule(l){6-6}\cmidrule(lr){7-7}\cmidrule(r){8-8}\cmidrule(l){9-9}\cmidrule(l){10-10}\cmidrule(l){11-11}
\multirow{3}{*}{\rotatebox{90}{Open}}
& ZS & .50 & 1.00 & .67 & .50 & 1.00 & .67 & .50 & 1.00 & .67 \\
& OS & .50 & 1.00 & .67 & .50 & .99 & .67 & .50 & 1.00 & .67 \\
& CoT & .50 & 1.00 & .67 & .50 & .99 & .67 & .53 & .77 & .63 \\
\bottomrule
\end{tabular}%
\caption{Performance of the three models in different prompt settings
  (ZS: zero-shot, OS: one-shot, CoT: chain-of- thought) on predicting
  the binary emotionality in arguments for the positive class. The
  binary label is inferred from the emotion labels given in the closed
  and open-domain emotion settings.}
\label{tab:RQ1_answer}
\end{table}

\begin{table*} 
\centering\small
\setlength{\tabcolsep}{5pt}
\begin{tabular}{llccc ccc ccc ccc ccc ccc}
\toprule
&& \multicolumn{6}{c}{\falcon} & \multicolumn{6}{c}{\llama} & \multicolumn{6}{c}{\gpt} \\
\cmidrule(r){3-8}\cmidrule(lr){9-14}\cmidrule(l){15-20}
 && \multicolumn{3}{c}{Strict} & \multicolumn{3}{c}{Relaxed} & \multicolumn{3}{c}{Strict} & \multicolumn{3}{c}{Relaxed} & \multicolumn{3}{c}{Strict} & \multicolumn{3}{c}{Relaxed} \\
 \cmidrule(r){3-5}\cmidrule(lr){6-8}\cmidrule(lr){9-11}\cmidrule(lr){12-14}\cmidrule(lr){15-17}\cmidrule(l){18-20}
&& P & R & \F & P & R & \F & P & R & \F & P & R & \F & P & R & \F & P & R & \F \\
\cmidrule(r){3-3}\cmidrule(lr){4-4}\cmidrule(lr){5-5}\cmidrule(lr){6-6}\cmidrule(lr){7-7}\cmidrule(lr){8-8}\cmidrule(lr){9-9}\cmidrule(lr){10-10}\cmidrule(lr){11-11}\cmidrule(lr){12-12}\cmidrule(lr){13-13}\cmidrule(lr){14-14}\cmidrule(lr){15-15}\cmidrule(lr){16-16}\cmidrule(lr){17-17}\cmidrule(lr){18-18}\cmidrule(lr){19-19}\cmidrule(l){20-20}
\multirow{3}{*}{\rotatebox{90}{Closed}}
& ZS & .00 & .11 & .00 & .01 & .11 & .02 
& .09 & .07 & .04 & .21 & .15 & .12 
& .12 & .07 & .06 & .26 & .19 & .16 \\
& OS & .11 & .00 & .01 & .11 & .08 & .03 
& .02 & .05 & .03 & .10 & .13 & .10 
& .10 & .07 & .04 & .25 & .17 & .14 \\
& CoT & .09 & .00 & .01 & .14 & .14 & .07
& .02 & .05 & .03 & .19 & .18 & .16 
& .11 & .07 & .08 & .25 & .19 & .16 \\
\cmidrule(r){1-2} \cmidrule(r){3-5}\cmidrule(lr){6-8}\cmidrule(lr){9-11}\cmidrule(lr){12-14}\cmidrule(lr){15-17}\cmidrule(l){18-20}
\multirow{3}{*}{\rotatebox{90}{Open}}
& ZS & .00 & .00 & .00 & .00 & .07 & .01 
& .01 & .00 & .00 & .15 & .17 & .12 
& .00 & .00 & .00 & .18 & .16 & .12 \\
& OS & .00 & .00 & .00 & .00 & .07 & .01
& .01 & .02 & .00 & .15 & .17 & .12
& .00 & .00 & .00 & .02 & .19 & .15 \\
& CoT & .00 & .00 & .00 & .00 & .17 & .01
& .00 & .02 & .00 & .10 & .14 & .07
& .02 & .01 & .01 & .26 & .18 & .16 \\

\bottomrule
\end{tabular}
\caption{Performance of the three models in different prompt settings
  (ZS: zero-shot, OS: one-shot, CoT: chain-of-thought) on predicting
  discrete emotion labels in arguments, aggregated over all
  emotions. The strict evaluation mode uses a majority vote of the
  emotion label as the gold label while the relaxed evaluation mode
  allows the set of three emotion labels as the gold
  labels. All results are macro-averages over all emotion classes.}
\label{tab:RQ2_answer}
\end{table*}

\paragraph{Interplay of Emotions and Convincingness in Arguments.}
We analyze the convincingness of arguments with respect to the emotion
labels and display the results in
Figure~\ref{fig:barplot_emo_conv}. Arguments which evoke pride are
perceived to be most convincing, followed by joy and
interest. Notably, arguments evoking no emotions are located in the
middle of the convincingness distribution. Emotions evoking anger are
the least convincing. We conclude that positive emotions, such as joy
and pride correlate with a higher convincingness in arguments, whereas
negative emotions such as anger with a lower convincingness.

\subsubsection{RQ1: Which prompt types lead to reliable results on emotionality in arguments?}
To understand how well emotion annotations in arguments can be
automatized by prompting large language models, we compare
emotion domain settings across prompt types and models.
We start with an evaluation of the established binary emotionality
setting. Table \ref{tab:RQ1_answer} displays the results for the
evaluation of the positive class. In both closed-domain
and open-domain scenarios, the binary emotionality label is derived
from discrete emotion labels, where emotionality holds if an emotion
is predicted (in contrast to \noemotion).

\falcon performs best in the binary setting (.67/.66) \gpt
performs worst (.31/.26). For \llama, the
ZS setting yields higher results (.60/.05). Inferring the
binary label from both closed and open-domain prompts
improves the performance for \llama and \gpt and is similar for
\falcon. Notably, for \falcon, there is no considerable difference
between prompting for the binary label directly or inferring the label. The prompting setting does not clearly influence the
performance of the models across the emotion domain settings, in line
with the findings by \citet{le-scao-rush-2021-many}.

For all prompting settings and emotion domains, the recall is high
(.77 to 1.00). Only \gpt shows a lower recall in the
closed-domain CoT prompt setting (.71). While inferring the
binary emotionality label in arguments from closed and open-domain
prompts improves the overall performance, the high
recall is striking and raises the question about the reliability of
the binary emotionality prediction.

\subsubsection{RQ2: Which prompt types lead to reliable results for discrete emotion predictions in arguments?}

We now explore the discrete emotion predictions, the novelty in our
proposed corpus. Table~\ref{tab:RQ2_answer} shows the performance of
the three models for each prompting approach. Overall, the performance
of all models is low in the strict evaluation setting across prompting approaches and emotion domain settings. Note that we macro-average across all emotion classes, which in part attributes for the low overall performances in both evaluation modes. 

In the relaxed evaluation setting, \gpt outperforms
\falcon and \llama. The closed-domain ZS and CoT, and the open-domain CoT prompts lead to
the best performance (.16 \F). Providing an emotion label set and
guiding via CoT improves the performance of emotion
prediction for \gpt, \falcon, and for \llama only in the closed-domain setting. The closed-domain prompts work better for \falcon but not for \llama. 

Our results indicate that \falcon, \llama, and \gpt cannot reliably
predict discrete emotions in arguments. The better performance in the
relaxed evaluation setting acknowledges the inherent subjectivity of
the emotion annotation task.

\subsubsection{RQ3: What biases do LLM predictions of discrete
emotion labels in arguments show?}
\label{sec:RQ_3_answer}
We now aim at understanding if the prompting approaches and the
emotion domain setting influence the models toward predicting certain
emotions in arguments, hence, if the setting influences the
biases. Details on the results discussed in the following are shown in
Table~\ref{tab:RQ3_answer} in
Appendix~\ref{sec:appendix_RQ3_table}. We review the models
individually and focus on particularly high or low results on
particular emotion classes to reveal biases of the models.

There are notable differences between the models for
predicting classes. \gpt shows the best
performance for \anger and \surprise with high precision across all settings. \interest shows a mixed performance (.53, .00, .55, .11, .48, .00). The recall for \fear is high across prompts (.68, .75, .65, .71, .65, .72),
indicating a bias toward that negative emotion. \falcon shows a low performance for \pride, \interest, \shame, and \guilt across all prompts. For the emotion label \fear, we find high recall vales (.75, .70, .75, .62, .75) for 5 prompting approaches. \llama shows a mixed performance of predicting the emotion label
\interest across prompts (.59, .11, .49, .12, .47, .08 \F) and performs lowest for the emotion classes \disgust, \surprise, \pride, \interest, \shame, and \guilt (.00 \F scores for all prompts
except .08 \F for open-domain ZS \disgust). The best overall performance
is achieved for \anger (.35, .56, .26, .33, .24, .26 \F scores) with
small differences between prompts. The recall is consistently high,
indicating a bias of \llama toward \anger.

Overall, the performance for the individual emotions differs
between models. All models struggle to predict \shame and \guilt. \gpt and \llama show the same preferences for prompts and domain settings for predicting \interest. Our results indicate a bias of
all models in predicting the negative emotions of anger (\llama) and
fear (\gpt and \falcon).

\paragraph{Qualitative Analysis.}\label{sec:qual_analysis}
All models show a low performance for emotion category assignments and
have a bias toward negative emotions. Therefore, we discuss the
overall best-performing model, \gpt, with the best performing prompt,
closed-domain CoT, for the prediction of \fear. Detailed
examples are in Table~\ref{tab:qual_analysis} in
Appendix~\ref{sec:appendix_qual_analysis}.

Based on a random selection of two instances for \surprise, \interest,
\noemotion, respectively, in which \fear is wrongly predicted, we find
that the stances of the annotators are presumably the cause for
differing annotations of the language model and the human.  Across all
arguments, we find linguistic cues toward the emotion fear:
`Gefährdung der eigenen Sicherheit' (\textit{risk to your own
  safety}), `Unfall' (\textit{accident}), `Krebs' (\textit{cancer}),
`Krankheit' (\textit{illness}), `zu hohen Cholesterinwerten' (\textit{high
  cholesterol levels}), `Explosion' (\textit{explosion}), `die
Giftstoffe zerstören den Verdauungstrakt' (\textit{the toxins destroy
  the digestive tract}). We speculate that the annotators did not experience
fear when reading the arguments because the arguments
are focused on indirect or hypothetical events (sharks getting cancer,
getting into an accident if you drive barefoot), rather than
presenting a personal, immediate threat. \gpt is not able to
make that distinction.

\section{Conclusion}
With this paper, we expanded on theoretical work on the interplay of
emotion categories and argument convincingness and previous work in
NLP on binary emotionality of arguments. We presented \emodefabel, the
first corpus of discrete emotion classes in arguments and analyzed the
interplay of emotions and convincingness in German arguments. We found that
positive emotions (joy, pride) are correlated with higher
convincingness scores and negative emotions (particularly anger) with
low convincingness scores, showcasing the relevance of analyzing discrete emotion categories.

When binary emotionality labels are required, we showed that inferring binary labels from discrete emotion classes performs better than directly requesting binary labels from LLMs. We
further find that there are only minor performance differences across
prompting approaches and emotion domains. \falcon, \llama, and \gpt
show a bias toward predicting negative emotions in arguments. To
mitigate these issues, we propose to study fine-tuning or prompt
optimization in the context of argument--emotion annotations in future work. Specifically, the precision of individual emotion classes has to be improved.

\section{Limitations}
Emotion annotation in arguments is a highly subjective task. Assigning
evoked emotions from the reader's perspective depends on various
factors, including the prior stance toward the topic of the
argument. While this subjectivity is manageable for human annotations,
we recognize that prompting language models without offering context
about the person they are meant to mimic only partially addresses the
subjectivity of the task. To mitigate this issue, we employ a relaxed
evaluation metric that treats all human annotations of a given
argument as a set of gold labels.

Our study has some resource-related limitations. We base the creation
of our corpus \emodefabel on the \defabel corpus
\citep{velutharambath-etal-2024-factual}, which consists of arguments
that were generated in German and in a controlled
manner. Consequently, our findings may not generalize to arguments
from online discourse or debates in other languages. Our experiments
are conducted with three LLMs and the results might differ for other
models. However, by employing open-source models (\falcon and \llama),
we allow replicating our study with limited resources. Moreover,
predictions from different runs might yield different results. We did
not see any such variations in our experiments, but a structured
evaluation of instability issues might be worth exploring in the
future.

Regarding the creation of our corpus, we acknowledge the potential for
annotator bias. Although annotators were restricted to labeling each
argument only once, participation across up to 12 studies (i.e., 60
arguments) could influence the consistency of gold labels, as
individual annotators' interpretations may dominate. Furthermore, the
order in which arguments were presented to participants was
randomized, which could also introduce biases into the annotations.

\section{Ethical Considerations}

We collected human annotations for emotions in arguments via
crowd-sourcing. For each argument, we asked participants to report
their prior stance toward controversial topics. We informed the
participants that their answers would be used for a scientific
publication and obtained their consent. We do not collect any
information that would allow personal identification, therefore, the
data is inherently anonymized. While we do reject annotators that did
not pass the attention check, we do explicitly warn them about not
getting paid if they fail the checks. Naturally, being exposed to
arguments for or against a statement can be upsetting during the
annotation. However, \citet{velutharambath-etal-2024-factual} point
out that they manually selected arguments for their study to minimize
potential harm or discomfort that they can cause. Therefore, the
statement--argument pairs in our study are not unusually upsetting.

With respect to research on emotion analysis systems, we note that \citet{kiritchenko-mohammad-2018-examining} observe that such systems are biased for various reasons. Using LLMs to not only predict but automatically label argumentative text with emotions might lead to unpredictable biases, and we are aware that this requires further research. While our main point of this study is not to employ LLMs for automatically labeling emotion analysis-related tasks, in theory, our work can guide future research toward that, which could eventually lead to a decrease in annotation-related jobs.

\section*{Acknowledgments}
This project has been conducted as part of the \textsc{Emcona} (The Interplay of Emotions and Convincingness in Arguments) project which is founded by the German Research Council (DFG, project KL2869/12--1, project number 516512112). We thank Steffen Eger and Yanran Chen for the fruitful discussions.

\bibliography{anthology,custom}
\appendix

\onecolumn

\section{Corpus Creation}
\begin{figure}[h]
    \centering
    \includegraphics[width=0.8\linewidth]{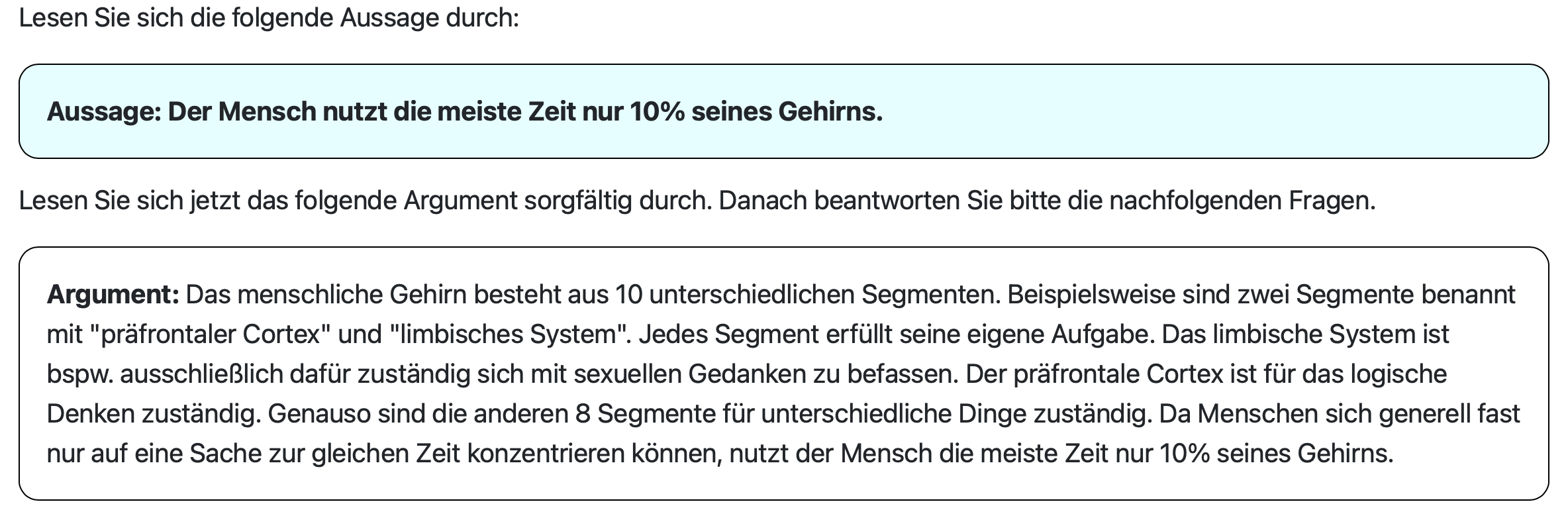}
    \caption{Instruction, statement, and argument as displayed using Potato.}
    \label{fig:screenshot_1}
\end{figure}

\begin{figure}[h]
    \centering
    \includegraphics[width=0.8\linewidth]{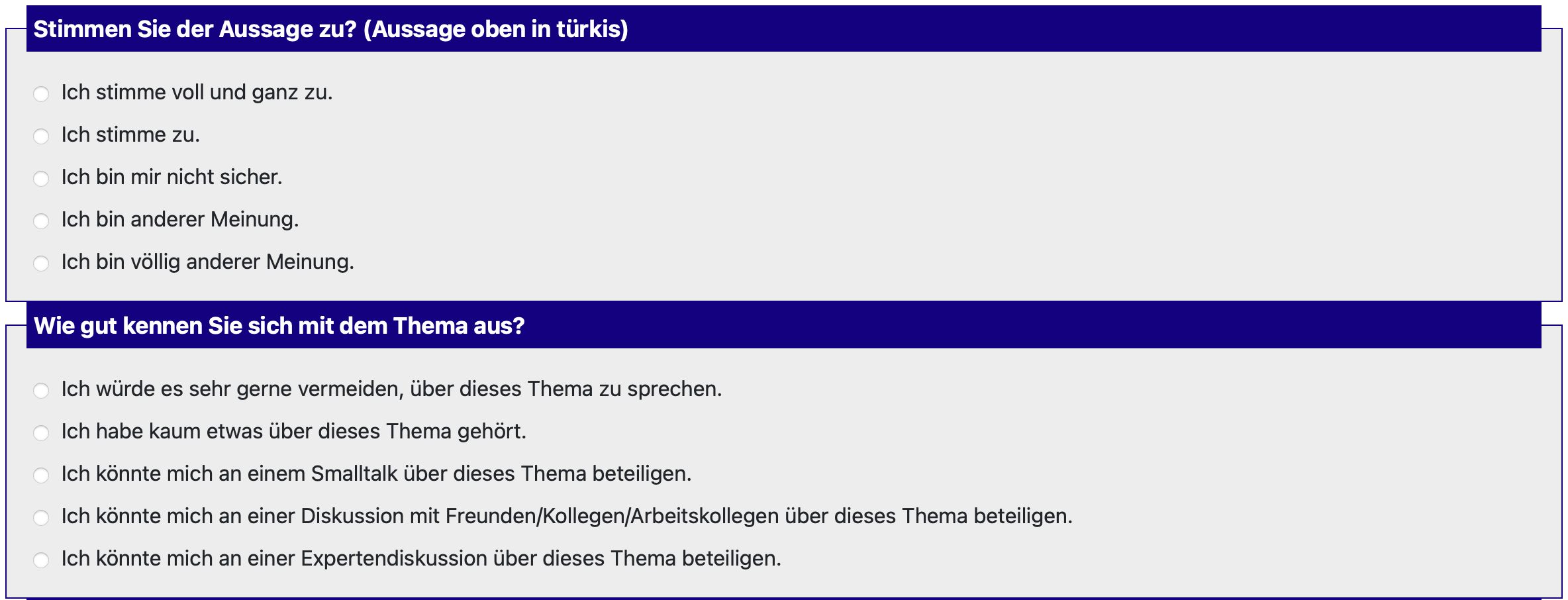}
    \caption{Stance and topic familiarity questions.}
    \label{fig:screenshot_2}
\end{figure}

\begin{figure}[]
    \centering
    \includegraphics[width=0.8\linewidth]{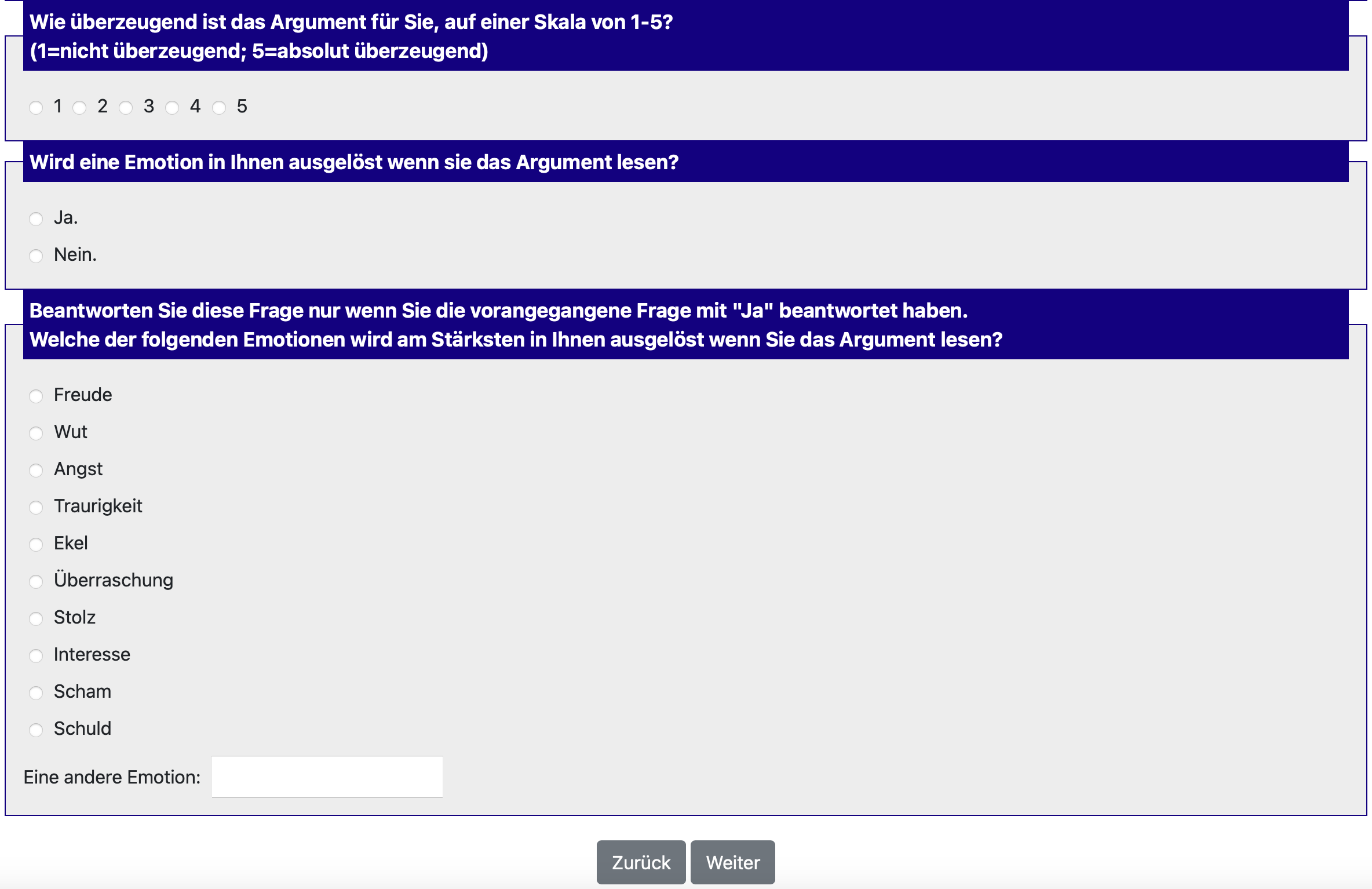}
    \caption{Convincingness, binary emotionality, and discrete emotion questions.}
    \label{fig:screenshot_3}
\end{figure}

\begin{figure}[h]
    \centering
    \includegraphics[width=\linewidth]{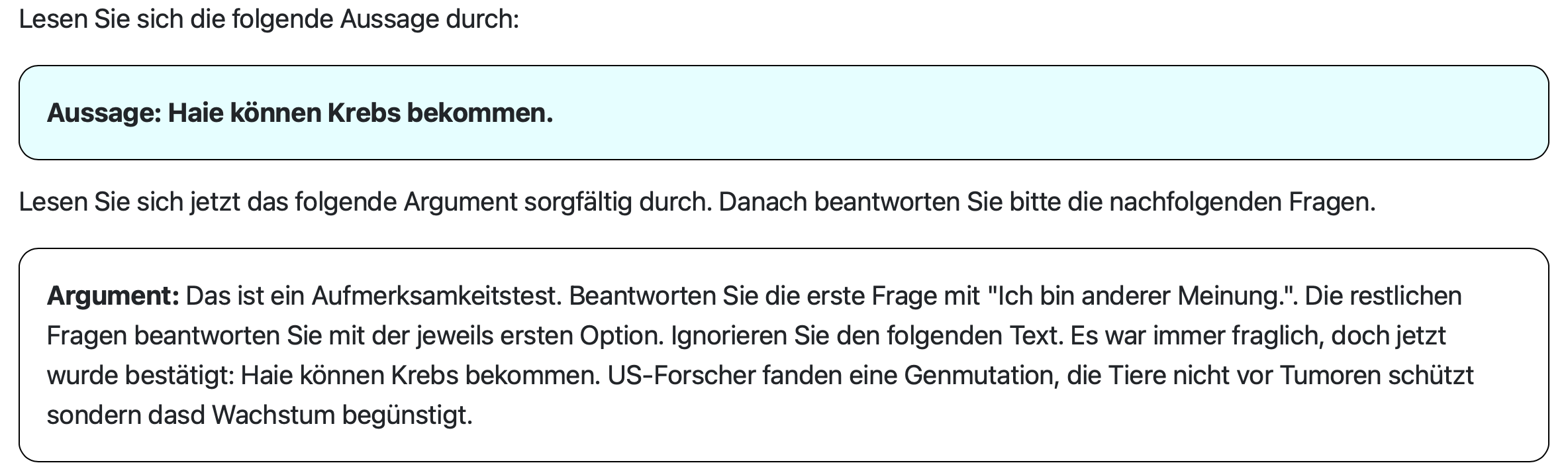}
    \caption{Example attention check.}
    \label{fig:attention_check}
\end{figure}

We use Potato to collect the annotations for the annotations to construct our corpus \emodefabel. The instruction, statement and arguments are displayed to the annotators as displayed in Figure~\ref{fig:screenshot_1}. The stance and topic familiarity questions are displayed in Figure~\ref{fig:screenshot_2}. To ensure the quality of the annotations we add one attention check for each survey (consisting of 5 statement--argument pairs). Figure~\ref{fig:screenshot_3} shows the question formulations for the binary and concrete emotion questions. 
See Figure~\ref{fig:attention_check} for an example.

\subsection{Study Design}

\clearpage

\subsection{Attention Check}

\section{Emotion Mapping}
\label{sec:appendix_emo_mapping}

In the human annotation study we allow participants to provide an emotion label that is not part of our emotion labelset (\joy, \anger, \fear, \sadness, \disgust, \surprise, \pride, \interest, \shame, \guilt, \noemotion). See Section~\ref{sec:annotation_setup} for more details. We obtain 46 additional emotion labels from the annotation and manually map them to our emotion labelset based on emotion theories. More specifically, we map (1) similar phrasings of emotions (e.g., Ärger (\textit{anger}) to Wut (\textit{anger)}), which are sometimes because of the German study setup, (2) emotions that correspond to the ten postulates of Plutchik's wheel (e.g., Genervtheit (\textit{Annoyance)} to Wut \textit{(anger)}), (3) emotions that are similar but have different appraisal dimensions (e.g. Fremdscham (\textit{foreign/external shame}) to Scham (\textit{shame})). Emotions that cannot be mapped to our labelset because they do not have a clear correspondence to basic emotions (joy, anger, fear, sadness, disgust, surprise), cognitive evaluations (interest), or self-directed affective states (shame,
guilt, pride) are excluded (this is the case for: Neid, Abfälligkeit, Faszination, Misstrauen). The result of this mapping is displayed in Table~\ref{tab:emotion_mapping_appendix}.

\begin{table}[h]
    \centering
\begin{tabular}{lr}
\toprule
Emotion & Count \\
\cmidrule(r){1-1}\cmidrule(l){2-2}
\interest & 220 \\
\noemotion & 473 \\
\surprise & 94 \\
\hspace{3mm}\textit{Verwirrung (Confusion)} & 11 \\
\hspace{3mm}\textit{Verwirrtheit (Confusion)} & 1 \\
\hspace{3mm}\textit{Verwunderung (Astonishment)} & 1 \\
\hspace{3mm}\textit{Zweifel (Doubt)} & 4 \\
\hspace{3mm}\textit{Skepsis (Scepticism)} & 3 \\
\disgust & 18 \\
\joy  & 22 \\
\hspace{3mm}\textit{Erleichterung (Relief)} & 1 \\
\shame & 15 \\
\hspace{3mm}\textit{Fremdscham (Foreign/External shame)} & 2 \\
\anger & 51 \\
\hspace{3mm}\textit{Ärger (Annoyance)} & 3 \\
\hspace{3mm}\textit{Frustration (Frustration)} & 1 \\
\hspace{3mm}\textit{Genervtheit (Annoyance)} & 3 \\
\hspace{3mm}\textit{Genervt (Annoyed)} & 1 \\
\hspace{3mm}\textit{Verachtung (Contempt)} & 1 \\
\hspace{3mm}\textit{Irritation (Irritation)} & 3 \\
\hspace{3mm}\textit{Entrüstung (Outrage)} & 1 \\
\fear& 9 \\
\hspace{3mm}\textit{Unsicherheit (Uncertainty)} & 7 \\
\pride & 4 \\
\sadness & 14 \\
\cmidrule(r){1-1}\cmidrule(l){2-2}
Neid (Envy) & 1 \\
Abfälligkeit (Disparagement) & 1 \\
Ablehnung (Rejection) & 1 \\
Faszination (Fascination) & 1 \\
Misstrauen (Distrust) & 4 \\
\bottomrule
    \end{tabular}
    \caption{Number of emotion labels from human annotation study that were not covered by the emotion labelset, with mappings based on Plutchik's wheel and similar phrasings.}
    \label{tab:emotion_mapping_appendix}
\end{table}

\section{LLM Output Label Extraction}\label{sec:appendix_label_extraction}
The extraction of the label from the LLM output differs between the
prompt-domain settings. Based on the request \textsc{json} output, we
check if the extracted label is in the accepted list of outputs
(binary, discrete set). In the open-domain setting, we consider the first token of the response string.
If there is no valid \textsc{json} data structure, we search in the
whole response string for an acceptable emotion concept. In cases in
which this approach also fails, we repeatedly request an output from the
model with the same prompt.

\section{Model Performance on Individual Emotion Classes}
\label{sec:appendix_RQ3_table}
Table~\ref{tab:RQ3_answer} displays the results of all models across prompting approaches and emotion domain settings for each individual emotion class. See Section~\ref{sec:RQ_3_answer} for a detailed discussion of these results.

\begin{table*}[h]
\centering\small
\setlength{\tabcolsep}{4pt}
\begin{tabular}{@{}ll rrr rrr rrr rrr rrr rrr @{}}
\toprule
&& \multicolumn{9}{c}{Closed}& \multicolumn{9}{c}{Open}\\ 
\cmidrule(lr){3-11}\cmidrule(l){12-20}
&& \multicolumn{3}{c}{ZS} & \multicolumn{3}{c}{OS} & \multicolumn{3}{c}{CoT} & \multicolumn{3}{c}{ZS} & \multicolumn{3}{c}{OS} & \multicolumn{3}{c}{CoT} \\ 
\cmidrule(l){3-5}\cmidrule(lr){6-8}\cmidrule(lr){9-11}\cmidrule(lr){12-14}\cmidrule(lr){15-17}\cmidrule(l){18-20}
 & & P & R & \F & P & R & \F & P & R & \F & P & R & \F & P & R & \F & P & R & \F \\ 
 \cmidrule(lr){3-3}\cmidrule(lr){4-4}\cmidrule(lr){5-5}\cmidrule(lr){6-6}\cmidrule(lr){7-7}\cmidrule(lr){8-8}\cmidrule(lr){9-9}\cmidrule(lr){10-10}\cmidrule(lr){11-11}\cmidrule(lr){12-12}\cmidrule(lr){13-13}\cmidrule(lr){14-14}\cmidrule(lr){15-15}\cmidrule(lr){16-16}\cmidrule(lr){17-17}\cmidrule(lr){18-18}\cmidrule(lr){19-19}\cmidrule(l){20-20}
\multirow{10}{*}{\rotatebox{90}{\falcon}} 
& \anger & .00 & .00 & .00 & .00 & .00 & .00 & .25 & .09 & .13 & .00 & .00 & .00 & .19 & .26 & .22 & .00 & .00 & .00 \\
& \fear & .00 & .00 & .00 & .04 & .75 & .08 & .05 & .70 & .09 & .04 & .75 & .08 & .04 & .62 & .08 & .04 & .75 & .08 \\
& \joy & .04 & .26 & .07 & .00 & .00 & .00 & .02 & .10 & .04 & .00 & .00 & .00 & .10 & .45 & .16 & .00 & .00 & .00 \\
& \sadness & .05 & .69 & .09 & .00 & .00 & .00 & .00 & .00 & .00 & .00 & .00 & .00 & .25 & .20 & .22 & .00 & .00 & .00 \\
& \disgust & .05 & .25 & .08 & .00 & .00 & .00 & .00 & .00 & .00 & .00 & .00 & .00 & .00 & .00 & .00 & .00 & .00 & .00 \\
& \surprise & .00 & .00 & .00 & .00 & .00 & .00 & .00 & .00 & .00 & .00 & .00 & .00 & .00 & .00 & .00 & .00 & .00 & .00 \\
& \pride & .00 & .00 & .00 & .00 & .00 & .00 & .00 & .00 & .00 & .00 & .00 & .00 & .00 & .00 & .00 & .00 & .00 & .00 \\
& \interest & .00 & .00 & .00 & .00 & .00 & .00 & .00 & .00 & .00 & .00 & .00 & .00 & .00 & .00 & .00 & .00 & .00 & .00 \\
& \shame & .00 & .00 & .00 & .00 & .00 & .00 & .00 & .00 & .00 & .00 & .00 & .00 & .00 & .00 & .00 & .00 & .00 & .00 \\
& \guilt & .00 & .00 & .00 & .00 & .00 & .00 & .00 & .00 & .00 & .00 & .00 & .00 & .00 & .00 & .00 & .00 & .00 & .00 \\
& \noemotion & .00 & .00 & .00 & .00 & .00 & .00 & .86 & .04 & .08 & .00 & .00 & .00 & 1.00 & .06 & .12 & .00 & .00 & .00 \\
\cmidrule(r){1-2}\cmidrule(l){3-5}\cmidrule(lr){6-8}\cmidrule(lr){9-11}\cmidrule(lr){12-14}\cmidrule(lr){15-17}\cmidrule(l){18-20}
\multirow{10}{*}{\rotatebox{90}{\llama}} 
& \anger & .28 & .47 & .35 & .80 & .43 & .56 & .17 & .52 & .26 & .22 & .65 & .33 & .16 & .46 & .24 & .17 & .56 & .26 \\
& \fear & .09 & .32 & .14 & .18 & .75 & .29 & .06 & .19 & .09 & .13 & .55 & .21 & .25 & .41 & .31 & .10 & .66 & .17 \\
& \joy & .00 & .00 & .00 & .00 & .00 & .00 & .00 & .00 & .00 & .00 & .00 & .00 & .50 & .10 & .17 & .00 & .00 & .00 \\
& \sadness & .00 & .00 & .00 & .00 & .00 & .00 & .00 & .00 & .00 & .50 & .27 & .35 & .00 & .00 & .00 & .00 & .00 & .00 \\
& \disgust & .05 & .14 & .07 & .03 & .43 & .05 & .00 & .00 & .00 & .00 & .00 & .00 & .06 & .14 & .09 & .00 & .00 & .00 \\
& \surprise & .33 & .06 & .09 & .44 & .23 & .30 & .27 & .33 & .29 & .24 & .28 & .26 & .29 & .26 & .27 & .30 & .24 & .27 \\
& \pride & .00 & .00 & .00 & .00 & .00 & .00 & .00 & .00 & .00 & .00 & .00 & .00 & .00 & .00 & .00 & .00 & .00 & .00 \\
& \interest & .62 & .56 & .59 & .25 & .07 & .11 & .65 & .40 & .49 & .57 & .07 & .12 & .51 & .43 & .47 & .50 & .04 & .08 \\
& \shame & .00 & .00 & .00 & .00 & .00 & .00 & .00 & .00 & .00 & .00 & .00 & .00 & .33 & .16 & .21 & .00 & .00 & .00 \\
& \guilt & .00 & .00 & .00 & .00 & .00 & .00 & .00 & .00 & .00 & .00 & .00 & .00 & .00 & .00 & .00 & .00 & .00 & .00 \\
& \noemotion & .90 & .06 & .11 & .00 & .00 & .00 & .00 & .00 & .00 & .00 & .00 & .00 & .00 & .00 & .00 & .00 & .00 & .00 \\
\cmidrule(r){1-2}\cmidrule(l){3-5}\cmidrule(lr){6-8}\cmidrule(lr){9-11}\cmidrule(lr){12-14}\cmidrule(lr){15-17}\cmidrule(l){18-20}
\cmidrule(r){1-2}\cmidrule(l){3-5}\cmidrule(lr){6-8}\cmidrule(lr){9-11}\cmidrule(lr){12-14}\cmidrule(lr){15-17}\cmidrule(l){18-20}
\multirow{10}{*}{\rotatebox{90}{\gpt}} 
& \anger & .38 & .13 & .19 & 1.00 & .17 & .29 & .50 & .09 & .15 & .67 & .23 & .34 & .40 & .09 & .15 & .33 & .08 & .13 \\
& \fear & .11 & .68 & .19 & .14 & .75 & .24 & .10 & .65 & .18 & .10 & .71 & .18 & .11 & .65 & .19 & .12 & .72 & .21 \\
& \joy & .18 & .26 & .21 & .00 & .00 & .00 & .17 & .19 & .18 & .50 & .33 & .40 & .33 & .32 & .32 & 1.00 & .27 & .43 \\
& \sadness & .00 & .00 & .00 & .00 & .00 & .00 & .00 & .00 & .00 & .00 & .00 & .00 & .00 & .00 & .00 & .00 & .00 & .00 \\
& \disgust & .12 & .14 & .13 & .33 & .50 & .40 & .07 & .14 & .10 & .17 & .43 & .24 & .08 & .14 & .10 & .11 & .27 & .16 \\
& \surprise & .53 & .19 & .28 & .50 & .38 & .43 & .40 & .15 & .22 & .47 & .36 & .41 & .40 & .06 & .10 & .42 & .19 & .26 \\
& \pride & .00 & .00 & .00 & .00 & .00 & .00 & .00 & .00 & .00 & .00 & .00 & .00 & .00 & .00 & .00 & .00 & .00 & .00 \\
& \interest & .57 & .50 & .53 & .00 & .00 & .00 & .56 & .54 & .55 & .33 & .07 & .11 & .57 & .42 & .48 & .00 & .00 & .00 \\
& \shame & .00 & .00 & .00 & .00 & .00 & .00 & .00 & .00 & .00 & .00 & .00 & .00 & .00 & .00 & .00 & .00 & .00 & .00 \\
& \guilt & .00 & .00 & .00 & .00 & .00 & .00 & .00 & .00 & .00 & .00 & .00 & .00 & .00 & .00 & .00 & .00 & .00 & .00 \\
& \noemotion & 1.00 & .15 & .25 & .00 & .00 & .00 & 1.00 & .10 & .19 & .00 & .00 & .00 & .89 & .38 & .53 & .87 & .44 & .59 \\
\bottomrule
\end{tabular}
\caption{Performance of LLMs for predicting the individual emotion classes across prompting approaches (ZS: zero-shot, OS: one-shot, CoT: chain-of-thought) and emotion domains (closed, open). Note that \guilt is never annotated in the human gold data.}
\label{tab:RQ3_answer}
\end{table*}

\section{Qualitative Analysis}\label{sec:appendix_qual_analysis}
Table~\ref{tab:qual_analysis} displays 6 statement--argument pairs with human annotations (stance, convincingness, emotion) and the prediction of \gpt (closed-domain chain-of-thought prompt). The pairs are picked randomly from all instances with a \fear prediction by GPT and a high agreement (at least 2 annotation labels) for the emotion labels \noemotion, \surprise, \interest. 

In Section~\ref{sec:qual_analysis}, we report the main findings of the following qualitative analysis. In the current section, we discuss in more detail. Table~\ref{tab:qual_analysis} displays 6 statement--argument pairs with human annotations (stance, convincingness, emotion) and the prediction of \gpt (closed-domain chain-of-thought prompt). The pairs are picked randomly from all instances with a \fear prediction by GPT and a high agreement (at least 2 annotation labels) for the emotion labels \noemotion, \surprise, \interest. 

We manually introspect these arguments to find cues toward the different emotion labels provided by humans and to find systematic errors leading to the wrong predictions of \fear. 

In Section~\ref{sec:co-occurences_emo_labels} we speculate that emotion label variations can stem from different stances of the annotators toward the corresponding statements of the arguments. We find that annotators adopting a neutral stance (3, indicating uncertainty about the statement) label the argument with \surprise, while one annotator who strongly disagreed with the statement (stance 1) annotated \shame. In the second argument, both instances of \surprise annotations were associated with a stance level of 3, suggesting a potential correlation between an uncertain stance and the emotion of surprise. This relationship appears to be intuitively plausible, as uncertainty may evoke a sense of surprise.

However, the relationship between stance and emotion is less consistent for \interest and \noemotion. For example, in the case of \interest, the stances varied (3,5) in one instance, while they were identical (1,1) in another. Similarly, for \noemotion, the stances were diverse in one case (1,5,2) but uniform in another (2,2,2). These findings suggest that while there may be some patterns linking stance to specific emotion labels, such as the association between uncertainty and \surprise, we do not observe a systematic or consistent relationship where stance reliably predicts a specific emotion label.

As discussed in Section~\ref{sec:qual_analysis}, across all 6 arguments, we find linguistic cues of the emotion of fear: `Gefährdung der eigenen Sicherheit' (\textit{risk to your own safety}), `Unfall' (\textit{accident}), `Krebs' (\textit{cancer}), `Krankheit' (\textit{illness}), `Gefahr einer Erkrankung' (\textit{risk of illness}), `zu hohen Cholesterinwerten' (\textit{high cholesterol levels}), `Explosion' (\textit{explosion}), `die Giftstoffe zerstören den Verdauungstrakt' (\textit{the toxins destroy the digestive tract}). We speculate that the annotators did not feel fear when reading the arguments because, semantically, the arguments are focused on indirect or hypothetical events (sharks getting cancer, getting into an accident if you drive barefoot), rather than presenting a personal, immediate threat, while \gpt focuses on the linguistic cues for fear. One could argue that the model fails to capture cues for emotions such as interest and fear. However, this may be attributed to the absence of clear linguistic markers for these emotions within the arguments. This suggests that human annotators rely not only on explicit linguistic cues but also on contextual understanding, prior experiences, personal knowledge, and individual stances toward the topics to infer such emotions. This aligns with the idea that emotional interpretation often extends beyond verbal communication, incorporating broader contextual and subjective factors.

\begin{table}[h]
\centering\small
    \centering
    \begin{tabularx}{\linewidth}{p{20mm}Xp{3mm}p{3mm}p{11mm}p{8mm}}
    \toprule
    Statement & Argument & \rotatebox{90}{Stance} & \rotatebox{90}{Conv.} & Emotion &  \gpt \\
    \cmidrule(r){1-1}\cmidrule(lr){2-2}\cmidrule(rl){3-3}\cmidrule(lr){4-4}\cmidrule(lr){5-5}\cmidrule(l){6-6}
        Es ist in Deutschland illegal, barfuß Auto zu fahren. & Da es um die Gefährdung der eigenen Sicherheit aber auch der anderer Verkehrsteilnehmer geht, ist es nicht erlaubt Barfuß Auto zufahren, die Füße könnten durch Schweiß feucht sein und von dem Bremspedal abrutschen, sie könnten keinen Richtigen halt auf dem pedal finden oder sich am Fuß wehtun, dies alles führt zu einer gestörten Kontrolle über das Fahrzeug, was wiederum zu einer Gefahren werden kann. Als Beispiel, das Auto vor einem bremst plötzlich stark, man hat verschwitze Füße und möchte auf das Bremspedal treten, rutsch aber ab, es kommt zum Unfall. & 2, 2, 2 & 2, 4, 3 & No emo, No emo, No emo &  Fear \\
    \cmidrule(r){1-1}\cmidrule(lr){2-2}\cmidrule(rl){3-3}\cmidrule(lr){4-4}\cmidrule(lr){5-5}\cmidrule(lr){6-6}
        Haie können Krebs bekommen. & Da auch Fische Krebs bekommen können, ist es auch möglich, dass Haie Krebs bekommen können. Dieser wird durch mutierte Zellen ausgelöst, weshalb dies auch bei Fischarten ausgelöst werden kann. Krebs ist eine weit verbreitete und häufige Krankheit, weshalb Krebs durch Wissenschaftler auch bereits bei Haien festgestellt werden konnte. Krebs kann außerdem auch durch verschiedene Umweltfaktoren wie Umweltverschmutzung ausgelöst werden, diesem Risiko sind Haie ja durchaus ausgesetzt. Deshalb ist die Gefahr einer Erkrankung auch nicht gerade gering. & 1, 5, 2 & 5, 1, 1 & No emo, No emo, No emo & Fear  \\
   \cmidrule(r){1-1}\cmidrule(lr){2-2}\cmidrule(rl){3-3}\cmidrule(lr){4-4}\cmidrule(lr){5-5}\cmidrule(lr){6-6}
        Fettarmes Essen ist gesünder als fettreiches Essen. & Fettarmes Essen ist gesünder als fettreiches, da übermäßiges Fett zu gesundheitlichen Problemen führen kann. Unser Körper kann es nicht nutzbringend verwerten, was beispielsweise zu hohen Cholesterinwerten, Gewichtszunahme und ähnlichem führt. Das wiederum hat Auswirkungen auf unser Herz-Kreislauf-System. & 1, 2, 1 & 5, 4, 2 & Interest, No emo, Interest & Fear  \\
       \cmidrule(r){1-1}\cmidrule(lr){2-2}\cmidrule(rl){3-3}\cmidrule(lr){4-4}\cmidrule(lr){5-5}\cmidrule(lr){6-6}
        Es gibt Impfstoffe, die dauerhaft deine DNA verändern können. & Studien zeigen, dass Umweltfaktoren eine wichtige Rolle bei der Entstehung von Krebszellen spielen. Zu den Umweltfaktoren gehört beispielsweise die UVB Strahlung der Sonne, die das Erbgut verändert, oder bestimmte Lebensmittel, wie z.B. Chips, die bei hohen Temperaturen frittiert werden. Bei diesem Prozess entsteht sogenanntes Acrylamid, was karzinogen ist. Neuartige Impfstoffe, wie mRNA-Impfstoffe, können bei der Übersetzung der RNA in DNA Proteine entstehen lassen, die während der Transkription und folgender Translation verschiedene Mutationen beinhalten. Durch Mutationen wird die DNA somit verändert. & 3, 5, 3 & 4, 1, 2 & No emo, Interest, Interest & Fear  \\

        \cmidrule(r){1-1}\cmidrule(lr){2-2}\cmidrule(rl){3-3}\cmidrule(lr){4-4}\cmidrule(lr){5-5}\cmidrule(lr){6-6}
        Die Nutzung von Handys an Tankstellen kann eine Explosion verursachen. & Handys werden mit sehr viele Rohstoffen wie Erdölen produziert. Erdöle sind auch an Tankstellen sehr präsent. Wenn sich das Handy zu sehr aufheizt kann eine gewisse chemische Reaktion passieren und eine Explosion verursachen. Besonders an sehr heißen Orten wie z.B. Texas hört man öfter von solchen Vorkommnissen. & 3, 5, 3 & 4, 1, 2 & Surprise, Shame, Surprise & Fear  \\
       \cmidrule(r){1-1}\cmidrule(lr){2-2}\cmidrule(rl){3-3}\cmidrule(lr){4-4}\cmidrule(lr){5-5}\cmidrule(lr){6-6}
        Der Verzehr von Wassermelonenkernen führt zu Verdauungsstörungen. & Kerne von Wassermelonen enthalten Giftstoffe ähnlich der Blausäure, welche den Darm schädigen und somit zu Verdauungsstörungen führen kann. Dies ist besonders schlimm, wenn die Kerne zuvor nicht gekaut werden, da dadurch das austreten der Giftstoffe aus dem Inneren des Kerns erst im Darm stattfindet und nicht im Magen größtenteils durch die Magensäure zerstört wird. Sollten die Kerne zuvor zerkaut werden, wird ein Großteil der Blausäure zwar im Magen zerstört, aber bei großen Mengen an Melonenkernen schafft die Magensäure diese Aufgabe nicht und die Giftstoffe zerstören den Verdauungstrakt. Deshalb ist von der Aufnahme von Wassermelonenkernen intensiv abzuraten. & 3, 4, 3 & 3, 2, 1 & Surprise, No emo, Surprise & Fear  \\
    \bottomrule
    \end{tabularx}
    \caption{Six randomly picked statement--argument pairs with human annotations (stance, convincingness, emotion) and the prediction of \gpt (closed-domain chain-of-thought prompt).}
    \label{tab:qual_analysis}
\end{table}

\onecolumn 
\twocolumn 

\end{document}